\documentclass[runningheads]{llncs}
\usepackage{graphicx}

\usepackage{tikz}
\usepackage{comment}
\usepackage{amsmath,amssymb} 
\usepackage{color}

\usepackage{ragged2e}
\usepackage{tabularx}

\usepackage[accsupp]{axessibility}  

\usepackage{xcolor}
\usepackage[linesnumbered,ruled,vlined]{algorithm2e}

\SetCommentSty{mycommfont}

\SetKwInput{KwInput}{Input}                
\SetKwInput{KwOutput}{Output}              

\usepackage{epsfig}
\usepackage{graphicx}
\usepackage{amsmath}
\usepackage{amssymb}
\DeclareMathOperator*{\argmax}{arg\,max}

\usepackage{dsfont}
\usepackage{amsfonts}
\usepackage[switch]{lineno}
\usepackage{enumitem}

\usepackage{float}

\usepackage{bbding}
\usepackage{booktabs}
\usepackage{multirow}
\usepackage{color}

\usepackage{wrapfig}

\begin{document}
\pagestyle{headings}
\mainmatter
\def\ECCVSubNumber{1005}  

\title{Neighborhood Collective Estimation for Noisy Label Identification and Correction} 


\titlerunning{Neighborhood Collective Estimation}
%
\author{Jichang Li\inst{1,2}\orcidID{0000-0001-5778-2232} \and
Guanbin Li\inst{1*}\orcidID{0000-0002-4805-0926} \and 
\\Feng Liu\inst{3}\orcidID{0000-0002-4811-7828} \and
Yizhou Yu\inst{2*}\orcidID{0000-0002-0470-5548}
\\
}
\authorrunning{J. Li et al.}
%
\institute{Sun Yat-sen University, Guangzhou 510006, China
\and
The University of Hong Kong, Hong Kong
\and
Deepwise AI Lab, Beijing, China\\
{\tt\small csjcli@connect.hku.hk, liguanbin@mail.sysu.edu.cn,\\  liufeng@deepwise.com, yizhouy@acm.org}
}

\maketitle

\newcommand\blfootnote[2]{%
\begingroup
\renewcommand\thefootnote{}\footnote{#1}%
\addtocounter{footnote}{-1}%
\endgroup
}

\blfootnote{*Corresponding Authors are Guanbin Li and Yizhou Yu.}

\begin{abstract}
Learning with noisy labels~(LNL) aims at designing strategies to improve model performance and generalization by mitigating the effects of model overfitting to noisy labels. The key success of LNL lies in identifying as many clean samples as possible from massive noisy data, while rectifying the wrongly assigned noisy labels. Recent advances employ the predicted label distributions of individual samples to perform noise verification and noisy label correction, easily giving rise to confirmation bias. To mitigate this issue, we propose Neighborhood Collective Estimation, in which the predictive reliability of a candidate sample is re-estimated by contrasting it against its feature-space nearest neighbors. Specifically, our method is divided into two steps: 1) Neighborhood Collective Noise Verification to separate all training samples into a clean or noisy subset, 2) Neighborhood Collective Label Correction to relabel noisy samples, and then auxiliary techniques are used to assist further model optimization. Extensive experiments on four commonly used benchmark datasets, i.e., CIFAR-10, CIFAR-100, Clothing-1M and Webvision-1.0, demonstrate that our proposed method considerably outperforms state-of-the-art methods.
\keywords{Learning with noisy labels, Neighborhood collective estimation, Confirmation bias.}
\end{abstract}

\section{Introduction}
Deep neural networks (DNNs) have achieved significant success in computer vision tasks, such as image classification~\cite{van2020survey,algan2021image,li2020deep,bendre2020learning,li2021cross,zhou2022generalized}, \textit{etc}. However, they rely heavily on tremendous quantities of high-quality manual annotations. To alleviate the need for extensive human annotations while improving the generalization capability of deep neural networks, learning with noisy labels (LNL) has been proposed to effectively leverage large-scale yet poorly-annotated datasets while mitigating the effects of model overfitting to noisy labels. 

To tackle the challenges imposed by LNL, previous works have proposed massive strategies~\cite{han2018coteaching,tanaka2018joint,li2020dividemix,ortego2021multi,yao2021jo}, including noisy label correction~\cite{arazo2019unsupervised,liu2020early}, noisy label or sample rejection\cite{li2020dividemix,yao2021jo,jiang2018mentornet,jiang2020beyond}, and noisy sample reweighing~\cite{wang2018iterative,ren2018learning,huang2020self}.  The mainstream pipeline first uses noise verification strategies to separate the original training set into a clean set and a noisy set, which contain training samples with clean labels and noisy labels respectively, in order to diminish the effect of noisy labels during model training. Then, (un)supervised learning or semi-supervised learning (SSL) based techniques are adopted to correct noisy labels and further optimize the classification model by regarding the clean set and noisy set as labeled and unlabeled samples respectively. In this scheme, original noisy labels are simply discarded for their high chances to be incorrect, avoiding the negative effect of noisy label memorization in the trained model.
 
\begin{figure}[t]
\centering
\includegraphics[width=3.0cm,height=2.5cm]{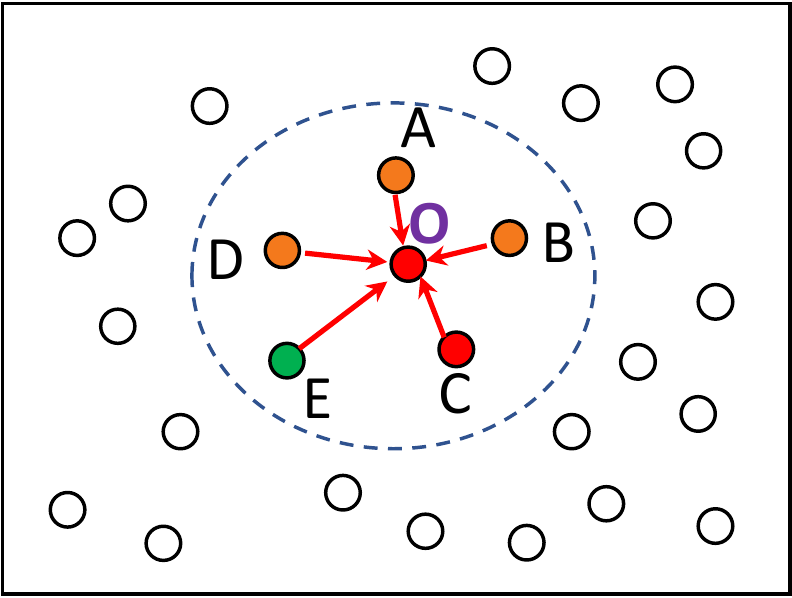}
\vspace{-0.3cm}
\caption{An illustration to exemplify our basic idea. Samples distributed within the dotted circle, including the candidate sample, Point O, and its nearest neighbors, \emph{i.e.}, Point A, B, C, D and E are close to each other in the feature-space neighborhood. Different colors indicate different labels~(either predicted label or given groundtruth label). In the noise verification stage, a given label of the candidate (Point O) is considered noisy if there is a huge inconsistency between the label distributions of the candidate and its nearest neighbors; and otherwise, the candidate is considered as a clean sample. Likewise, in the noise correction stage, a noisy sample discards the given noisy label and is relabeled through a neighborhood collective estimation process involving its contrastive neighbors
}
\label{fig1}
\end{figure}

In the context of learning with noisy labels, there may exist classes with imbalanced noisy or clean samples, especially in real-world noisy datasets such as Clothing-1M~\cite{Xiao2015Learning} and Webvision-1.0~\cite{Li2017a}. For instance, there might be a relatively high proportion of noisy labels in some hard-to-annotate classes; on the other hand, a trained model may produce low-confident predictions on a relatively high proportion of hard-to-learn clean samples in some classes, making existing noise identification algorithms incorrectly identify them as noisy samples. As a result, noise accumulation may take place implicitly in such classes, making the trained model produce unreliable label predictions. The above scenarios could make an LNL algorithm fall into the so-called confirmation bias~\cite{tarvainen2017mean,pseudoLabel2019}, which causes the algorithm to favor incorrect training labels that have been confirmed with predicted labels in earlier training iterations.  
In this context, relying too much on the potentially biased label predictions for individual training samples would increase the risk of incorrectly identifying noisy labels in the noise verification stage. Moreover, confirmation bias also exists in the subsequent noise correction stage, where SSL or other methods, such as label-guessing~\cite{li2020dividemix,nishi2021augmentation,zheltonozhskii2021contrast} and label re-assignment~\cite{yao2021jo}, construct pseudo-labels for unlabeled samples in the noisy set using potentially biased label predictions. Apparently, model training in the optimization stage would strengthen this bias as more confident but incorrect predictions would defy new changes, and subsequently even deteriorate model performance in high noise ratio scenarios.

We are inspired by the premise of contrastive learning that samples from the same class should have higher similarity in the feature space than those from different classes~\cite{NIPS2013_9aa42b31,oord2018representation,gunel2020supervised}. Therefore, we approach learning with noisy labels from a different perspective and propose Neighborhood Collective Estimation (NCE), in which we re-estimate the predictive reliability of a candidate sample by contrasting it against its feature-space nearest neighboring samples. Herein, we borrow the concept from contrastive learning, and then name such neighboring samples of the candidate as contrastive neighbors. Leveraging contrastive neighbors enriches the predictive information associated with the candidate and also makes such information relatively unbiased, thereby improving the accuracy of noisy label identification and correction. Fig.~\textcolor{red}{\ref{fig1}} displays the basic idea of the proposed method.

Specifically, to abide by the mainstream LNL pipeline, we divide our method into two steps: 1) Neighborhood Collective Noise Verification (NCNV) to separate all training samples into a clean set and a noisy set, 2) Neighborhood Collective Label Correction (NCLC) to relabel noisy samples. 
In the NCNV stage, a candidate sample is considered noisy when there is a huge inconsistency between the one-hot vector of the given 
label of the candidate and the label distributions of its contrastive neighbors predicted using the trained model.
In the NCLC stage, we only relabel noisy samples whose predicted label distribution is sufficiently similar to the given labels of neighboring clean samples, and the corrected label of a noisy sample is related to a weighted combination of the given labels of neighboring clean samples. 
Once we have identified clean samples and relabeled noisy ones, we leverage off-the-shelf and well-established techniques, such as mixup regularization~\cite{zhang2017mixup} and consistency regularization~\cite{sohn2020fixmatch}, to perform further SSL-based model training.

In summary, the main contributions are as follows.

\begin{itemize}
\item We propose Neighborhood Collective Estimation for learning with noisy labels, which leverages contrastive neighbors to obtain richer and relatively unbiased predictive information for candidate samples and thus mitigates confirmation bias.
\item Concretely, we design two steps called Neighborhood Collective Noise Verification and Neighborhood Collective Label Correction to identify clean samples and relabel noisy ones respectively.
\item We evaluate our method on four widely used LNL benchmark datasets, \emph{i.e.}, CIFAR-10~\cite{krizhevsky2009learning}, CIFAR-100~\cite{krizhevsky2009learning}, Clothing-1M~\cite{Xiao2015Learning} and Webvision-1.0~\cite{Li2017a}, and the results demonstrate that our proposed method considerably outperforms state-of-the-art LNL methods.
\end{itemize}

\section{Related Work}
In this section, we focus on noise verification and label correction that are means involved in current dominant pipeline to address the LNL problem. 

\subsection{Noise Verification}

Noise verification involves sample selection to choose and remove noisy labels within the training datasets. Proper noise verification strategies are necessary and several earlier works~\cite{han2018coteaching,yu2019does,jiang2018mentornet} have shown that samples with smaller cross-entropy loss are prone to hold clean labels, assuming that deep neural networks prefer to memorize simple patterns first rather than overfit to noisy labels. Also, some recently superior methods made efforts to model per-sample loss distributions with Beta Mixture Models (BMM)~\cite{ma2011bayesian} or Gaussian Mixture Models (GMM)~\cite{permuter2006study} to separate noisy labels from all the training samples~\cite{arazo2019unsupervised,li2020dividemix,nishi2021augmentation,zheltonozhskii2021contrast,huang2021learning,yang2022learning}. However, based on the predicted label distributions of individual candidate samples to identify the training samples, the above-stated noise verification strategies tend to fall into confirmation bias. Previous works have also attempted to identify noisy labels by leveraging neighborhood information. They either use neighborhood samples to remove noisy labels or re-weight them~\cite{wu2020topological,deepknn2020,ortego2021multi,zhu2021clusterability,2021NGC}. For example, Bahri \emph{et al.}~\cite{deepknn2020} proposed to identify noisy label by searching nearest neighbors based on the model predictions of a KNN classifier, while Zhu \emph{et al.}~\cite{zhu2021clusterability} uses feature-space neighbors to help estimate a noise transition matrix. 
In our work, we employ neighborhood collective estimation to realize both the identification and correction of noise labels, and make the two promote each other, to achieve better noise label learning.

\subsection{Label Correction}
To alleviate the effect of noisy memorization, noisy labels are discarded simply, and then label correction is adopted to relabel unlabeled samples~\cite{Dimensionality2018ma,song2019selfie,li2020dividemix,nishi2021augmentation,yao2021jo,zheltonozhskii2021contrast}. This aims to give reliable pseudo-labels and support subsequent model training so as to achieve better performance. For example, ``SELFIE'' proposed by Song \emph{et al.}~\cite{song2019selfie} tried to perform label correction by considering model predictions from past selecting clean labels. Also, Li \emph{et al.}~\cite{li2020dividemix} ``co-guessed'' pseudo-labels for unlabeled (noisy) samples via ensembling predictions of coupled networks, while Yao \emph{et al.}~\cite{yao2021jo} employed label re-assignment to provide pseudo-labels with the predictions of a temporally averaged model. Different from those as mentioned above, we correct noisy labels with the aid of neighboring labeled samples. This can relatively avoid confirmation bias that derives from model predictions at individual samples.

\section{The Proposed Method}

\begin{figure*}[t]
    
\centering
\includegraphics[width=9cm,height=4.0cm]{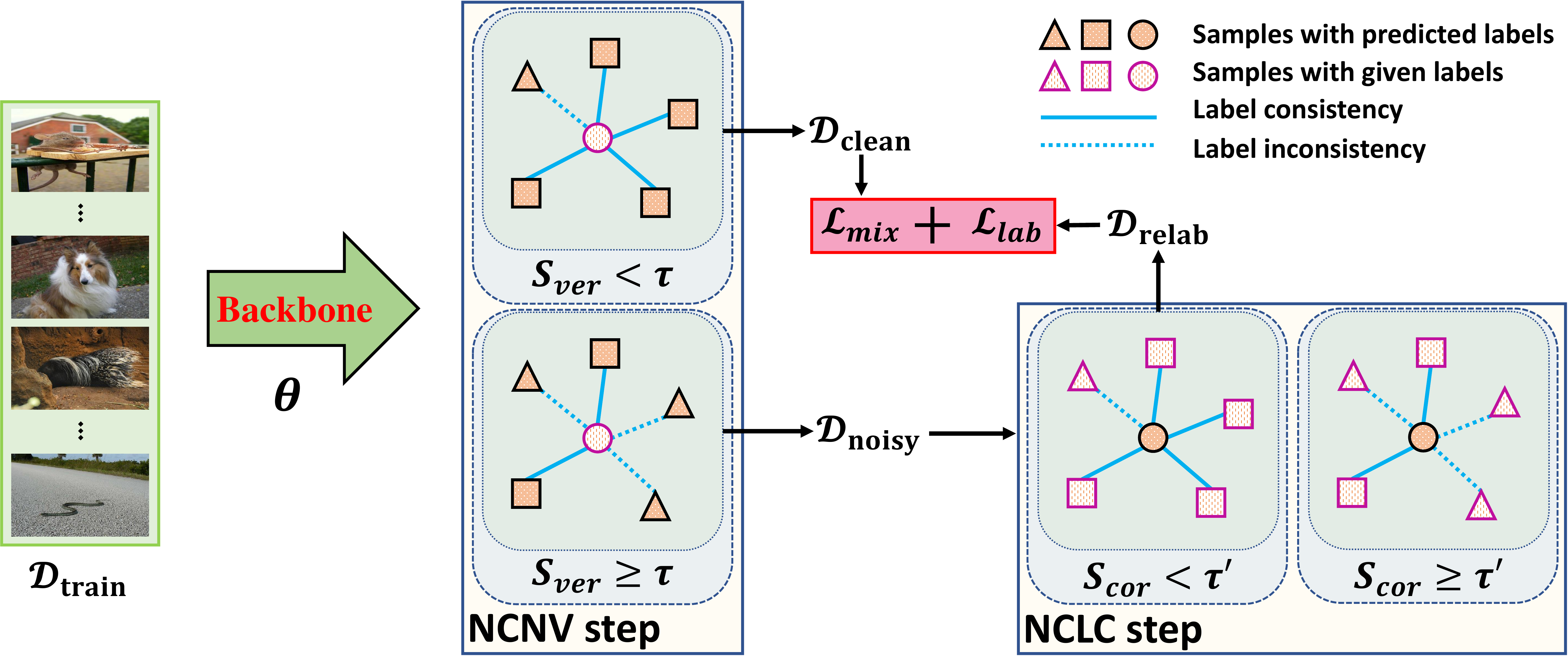}
\vspace{-0.3cm}     
\caption{Our proposed steps for learning with noisy labels. Triangles and squares represent contrastive neighbors from two different classes while circles denote the candidate samples in various steps. We assume the candidates belong to the class represented by the squares. In this work, we design two steps called Neighborhood Collective Noise Verification (NCNV) and Neighborhood Collective Label Correction (NCLC) to identify clean samples and relabel noisy ones respectively. Both steps leverage contrastive neighbors to obtain richer and relatively unbiased predictive information for candidate samples and thus mitigate confirmation bias} 
\label{fig2}
\vspace{-0.5cm}
\end{figure*}

{\bf Problem formulation.} 
Learning with noisy labels seeks an optimal model trained with a large-scale noisy dataset $\mathcal{D}_{{\bf train}}=\{\left(x_i, y_i\right)\}_{i=1}^{N}$, where $N$ is the number of sample-label pairs and each pair consists of a training sample $x_i$ and its associated label $y_i$ over $C$ classes while whether the given label is noisy or clean is unknown. During the training process, a sample is fed into a model being trained, that is parameterized by $\theta$ and contains a feature extractor $\Phi$ and a classifier with a softmax layer, to obtain its corresponding feature representation $\Phi(x_i)$ and class probabilities $p(y|x_i)$ respectively.

{\bf Contrastive neighbors.} We contrast a candidate sample against its feature-space nearest neighbors to enrich and diversify predictive information of the candidate. Such nearest neighbors are called contrastive neighbors in this paper. First, to compute feature similarity between a candidate sample $x_i$ and one of its feature-space neighbors $x_j$, we define a similarity function:
\begin{equation}
\label{featsimi}
d(x_i, x_j) = \frac{\Phi(x_i)^{\top}\Phi(x_j)}{\lVert \Phi(x_i) \rVert\lVert \Phi(x_j) \rVert} ,
\end{equation}
where $d\left(\cdot, \cdot\right)$ denotes the cosine distance metric. Then, we set up a pairwise connection between the two samples and quantify the discrepancy between their label distributions through the Jensen-Shannon (JS) divergence as follows,
\begin{equation}
J(p_i, p_j) = \frac{1}{2} KL(p_i||\frac{p_i+p_j}{2}) + \frac{1}{2} KL(p_j||\frac{p_i+p_j}{2}),
\end{equation}
where $KL(\cdot||\cdot)$ represents the Kullback-Leibler (KL) divergence, and for sample $x_i$ ( or $x_j$), in different contexts, $p_i$ (or $p_j$) represents either its probabilistic label distribution predicted using a trained model or its given ground-truth label. $J(\cdot, \cdot)$ returns values in the range of [0,1], and the use of JS divergence allows us to measure the discrepancy between the probabilistic label distributions of different samples. $J(p_i, p_j) \rightarrow 0$ indicates that the label distributions of $p_i$ and $p_j$ are very similar while $J(p_i, p_j) \rightarrow 1$ means the label distributions of these two samples are of great difference.

{\bf Overview.} 
In this paper, we propose Neighborhood Collective Estimation (NCE) to tackle learning with noisy labels. In detail, we first propose Neighborhood Collective Noise Verification (NCNV) to identify noisy labels in $\mathcal{D}_{{\bf train}}$ and divide $\mathcal{D}_{{\bf train}}$ into clean subset $\mathcal{D}_{{\bf clean}}$ and noisy subset $\mathcal{D}_{{\bf noisy}}$. Then, we propose Neighborhood Collective Label Correction (NCLC) to relabel selected samples from $\mathcal{D}_{{\bf noisy}}$ and form a new subset $\mathcal{D}_{{\bf relab}}$. Finally, we leverage auxiliary techniques to perform model fine-tuning so as to further optimize our model. The diagram and the training procedure of our proposed model have been summarized in Fig.~\textcolor{red}{\ref{fig2}} and Algorithm~\textcolor{red}{\ref{pseudocode}}, respectively.

\begin{algorithm}[H]

\DontPrintSemicolon
  
  \KwInput{Dataset $\mathcal{D}_{{\bf train}}$; Number of training epochs $T_{tr}$; Number of warm-up epochs $T_{wu}$; Learning rate $\eta$}
  \KwOutput{Optimal model parameter $\theta$}

   \For{$t \rightarrow 1 \cdots T_{tr}$}
   {    
      \uIf{$t < T_{wu}$}{
            \tcc{The warm-up step.}
       		WarmUp($\mathcal{D}_{{\bf train}}$; $\theta$). 
       		\tcp*{Initialize the model with a "WarmUp" function.}
      }
      \Else{
           \tcc{The NCNV step.}
 
            Use Eq. (\textcolor{red}{\ref{labelsplit}}) to split $\mathcal{D}_{{\bf train}}$ into clean samples $\mathcal{D}_{{\bf clean}}$ and noisy ones $\mathcal{D}_{{\bf noisy}}$.
            
            \tcc{The NCLC step.}
            
            Use Eq. (\textcolor{red}{\ref{relabel}}) to relabel a subset of samples from $\mathcal{D}_{{\bf noisy}}$ and form a new subset $\mathcal{D}_{{\bf relab}}$.
            
            \tcc{The model fine-tuning step.}
            
            Randomly sample mini-batches from $\mathcal{D}_{{\bf clean}}$ and $\mathcal{D}_{{\bf relab}}$.
            
            Update model parameter $\theta$ by applying SGD with $\eta$ to Eq. (\textcolor{red}{\ref{finalloss}}).
      }
   }
\caption{Learning with Noisy Labels based on Neighborhood Collective Estimation}
\label{pseudocode}
\end{algorithm}

\subsection{Neighborhood Collective Noise Verification}
In an effort to identify label noise for the task of LNL, most recent research establish sample selection criteria on the basis of predicted label distributions of individual samples~\cite{han2018coteaching,yu2019does,jiang2018mentornet,arazo2019unsupervised,li2020dividemix}, thus it is hard for them to avoid confirmation bias. Aiming at mitigating such bias, we formulate a novel noise verification function that determines whether a candidate is a noisy sample or not through the estimation of its label inconsistency score, which measures the degree of inconsistency between the label distributions of the candidate sample and its contrastive neighbors. 
Specifically, given a candidate sample-label pair $\left(x^{(c)}, y^{(c)}\right)\in\mathcal{D}_{{\bf train}}$, we first find its $K$ nearest neighbors in the feature space using the cosine similarity in Eq. (\textcolor{red}{\ref{featsimi}}) and then declare them as contrastive neighbors, as formulated below.
\begin{equation}
\{{x^{(c)}_{k}}\}, k=1,\cdots,K\leftarrow {\bf KNN}(x^{(c)}; \mathcal{D}_{{\bf train}}; K),
\end{equation}
where ${\bf KNN}(x^{(c)}; \mathcal{D}_{{\bf train}}; K)$ is a function that returns $K$ most similar samples in $\mathcal{D}_{{\bf train}}$ for the candidate sample $x^{(c)}$. Note that $x^{(c)}$ is temporarily removed from $\mathcal{D}_{{\bf train}}$ at this moment.

Then, the neighborhood-based label inconsistency score for the given label of the candidate can be defined as follows,
\begin{equation}
\label{con_sco}
S_{ver}(x^{(c)}, y^{(c)}) = \frac{1}{K}\sum_{k=1}^K J(p_y(y^{(c)}), p(y|x^{(c)}_{k})),
\end{equation}
where $p_y(y^{(c)})$ is the one-hot vector for the given ground-truth label $y^{(c)}$ of the candidate sample and $p(y|x^{(c)}_{k})$ stands for the probabilistic label distribution of the $k$-th contrastive neighbor predicted using a classification model trained with all original samples including both clean and noisy ones. Here, instead of the model prediction at the candidate sample, we make use of model predictions at its contrastive neighbors, implicitly diversifying the predictive information of the candidate sample and making it relatively unbiased.

After computing the label inconsistency score for every candidate sample, we observe that if the given ground-truth label of a candidate sample is significantly different from the model prediction of its contrastive neighbor samples,  \emph{i.e.},  of large inconsistency, then the given label is very likely to be a noisy label.
Therefore, by setting a threshold $\tau$, we can classify candidate sample $x^{(c)}$ as a noisy sample if $S_{ver}(x^{(c)}, y^{(c)})\ge\tau$, and otherwise, a clean one. To this end, we can obtain $\mathcal{D}_{{\bf clean}}$ and $\mathcal{D}_{{\bf noisy}}$ as follows,
\begin{equation}
\label{labelsplit}
\begin{split}
    \mathcal{D}_{{\bf clean}} \leftarrow \{\left(x_i, y_i\right) | S_{ver}(x_i, y_i)<\tau, \forall \left(x_i, y_i\right)\in\mathcal{D}_{{\bf train}}\}, \\
    \mathcal{D}_{{\bf noisy}} \leftarrow \{\left(x_i, \right) | S_{ver}(x_i, y_i)\ge\tau, \forall \left(x_i, y_i\right)\in\mathcal{D}_{{\bf train}}\}.
\end{split}
\end{equation}

\subsection{Neighborhood Collective Label Correction}
After the neighborhood collective noise verification (NCNV) stage, we treat samples from $\mathcal{D}_{{\bf clean}}$ and $\mathcal{D}_{{\bf noisy}}$ as labeled and unlabeled samples respectively by simply discarding noisy labels to prevent noise memorization in the resulted classification model. To leverage the unlabeled samples, some studies have taken pseudo-labeling based methods to mine discriminative cues for model training~\cite{li2020dividemix,yao2021jo,ortego2021multi}, yet all of them resort to model predictions at individual unlabeled samples, again tracing back to the unavoidable bias. On the contrary, we set up neighborhood collective label correction~(NCLC) stage, which corrects noisy labels by relying on neighboring clean samples to obtain more reliable and relatively unbiased pseudo-labels. 

As in the NCNV stage, we first find $K$ contrastive neighbors for each noisy sample $x^{(u)}\in\mathcal{D}_{{\bf noisy}}$ according to the ranked feature similarities between $x^{(u)}$ and its neighbors, as formulated below. At this time, we require all its contrastive neighbors to belong to the clean set $\mathcal{D}_{{\bf clean}}$.
\begin{equation}
\{{(x^{(u)}_{k}, y^{(u)}_{k})}\}, k=1,\cdots,K\leftarrow {\bf KNN}(x^{(u)}; \mathcal{D}_{{\bf clean}}; K),
\end{equation}
where $(x^{(u)}_{k}, y^{(u)}_{k})$ is a sample-label pair from $\mathcal{D}_{{\bf clean}}$. Unlike the NCNV stage, the ground-truth label information of contrastive neighbors is required in this stage. 

Afterwards, we perform the following label consistency check between each candidate sample and its contrastive neighbors to mine those noisy samples that are similar to their neighboring samples in both the feature and label space,
\begin{equation}
\label{scor}
S_{cor}(x^{(u)}) = \frac{1}{K}\sum_{k=1}^K J(p(y|x^{(u)}), p_y(y^{(u)}_{k})),
\end{equation}
where $J(p(y|x^{(u)}), p_y(y^{(u)}_{k}))$ computes the discrepancy between the probabilistic label distribution of the candidate sample $x^{(u)}$ predicted using the trained classification model,
and the one-hot vector for the given ground-truth label of its $k$-th contrastive neighbor. A large $S_{cor}(x^{(u)})$ indicates that the predicted label of the candidate sample is highly dissimilar to the clean and definite labels of its contrastive neighbors, suggesting that the candidate sample may lie near the decision boundary of the model. To be safe, we drop such candidate noisy samples if $S_{cor}(x^{(u)}) \ge \tau^{\prime}$, where a second threshold $\tau^{\prime}$ is used. In contrast, a candidate sample that satisfies $S_{cor}(x^{(u)})<\tau^{\prime}$ is more likely to be farther away from the decision boundary and could derive a more reliable pseudo-label from its contrastive neighbors. Therefore, we define a label correction function to generate a new label for such a noisy sample as follows,
\begin{equation}
 {\bf Correct}(x^{(u)}) = \argmax_c \sum_{k=1}^K w(x^{(u)}; k) \cdot p_y(y^{(u)}_{k}),
\end{equation}
where we use $w(x^{(u)}; k)=1 - J(p(y|x^{(u)}), p_y(y^{(u)}_{k}))$ to approximate the probability that the candidate sample belongs to the same class as its $k$-th contrastive neighbor, and $c=1,\cdots,C$ indicates the $c$-th component of a label distribution vector has the maximum value.
For convenience, we set $\hat{y}^{(u)}= {\bf Correct}(x^{(u)})$.

Finally, we define a new sample collection that contains all relabeled noisy samples as follows,
\begin{equation}
\label{relabel}
\begin{split}
    \mathcal{D}_{{\bf relab}} \leftarrow  \{\left(x_i, \hat{y}_{i}\right) |\hat{y}_{i}= {\bf Correct}(x_i), S_{cor}(x_i)<\tau^{\prime}, \forall x_i\in \mathcal{D}_{{\bf noisy}}\}.
\end{split}
\end{equation}

\subsection{Training Objectives}
Once we have the clean set $\mathcal{D}_{{\bf clean}}$ and relabeled set $\mathcal{D}_{{\bf relab}}$ respectively from the NCNV and NCLC steps, we use both datasets together to further optimize the classification model through fine-tuning. Auxiliary techniques are incorporated during model optimization. Since the initial classification model trained using both clean and noisy samples memorizes noisy labels during its training process and Mixup~\cite{zhang2017mixup} can effectively attenuate such noise memorization, we first employ the mixup regularization to construct augmented samples through linear combinations of existing samples from $\mathcal{D}_{{\bf clean}}$. 

Given two existing samples $(x_i, y_i)$ and $(x_j, y_j)$ from $\mathcal{D}_{{\bf clean}}$, an augmented sample $(\tilde{x}, \tilde{y})$ can be generated as follows,
\begin{equation}
    {\tilde{x}}={{\lambda}}{x_i}+({1-{\lambda}}){x_j}, {\tilde{y}}={{\lambda}}{p_y(y_i)}+({1-{\lambda}}){p_y(y_j)},
\end{equation}
where ${\lambda}\sim Beta(\alpha)$ is a mixup ratio and $\alpha$ is a scalar parameter of Beta distribution. The cross-entropy loss applied to $B$ augmented samples in each mini-batch is defined as follows,
\begin{equation}
{\mathcal{L}}^{mix}=-\sum_{b=1}^{B}{\tilde{y}_b \log{p(y|\tilde{x}_b)}}.
\end{equation}

In the NCLC stage, more reliable pseudo-labels are assigned to noisy samples farther away from the decision boundary. To leverage these relabeled samples during model optimization, we apply consistency regularization to them to further enhance the robustness of the model~\cite{englesson2021consistency}. Label consistency is a good choice to achieve this goal because it encourages the fine-tuned model to produce the same output when there are minor perturbations in the input~\cite{sohn2020fixmatch}. In practice, we enforce label consistency through the following loss:
\begin{equation}
\label{labloss}
{\mathcal{L}}^{lab}=-\sum_{{b^{\prime}}=1}^{B^{\prime}}{p_y(y_{b^{\prime}}) \log{p(y|{\bf Aug}(x_{b^{\prime}}))}},
\end{equation}
where $B^{\prime}$ relabeled samples $(x_{b^{\prime}}, y_{b^{\prime}})\in\mathcal{D}_{{\bf relab}}$ are chosen in each iteration, $p_y(y_{b^{\prime}})$ is the one-hot vector of the pseudo-label of $x_{b^{\prime}}$, ${\bf Aug}(\cdot)$ denotes the function that perturbs the chosen samples using Autoaugment technique proposed in~\cite{cubuk2019autoaugment}, and $p(y|{\bf Aug}(x_{b^{\prime}}))$ is the predicted label distribution of the perturbed sample. Proved by our experiments, this label consistency loss can be also applied to the selected clean samples from $\mathcal{D}_{{\bf clean}}$, especially under low noise ratios, to better boost the performance of the model.

As stated above, the overall loss function for final model fine-tuning is a combination of the cross-entropy and label consistency losses,
\begin{equation}
\label{finalloss}
{\mathcal{L}}^{overall}={\mathcal{L}}^{mix}+ \gamma{\mathcal{L}}^{lab},
\end{equation}
where $\gamma$ is a trade-off scalar to balance those two loss terms.

\section{Experiments}
\subsection{Experimental Setup}
{\bf Implementation.} We highlight the effectiveness of our proposed NCE method on four standard LNL benchmark datasets: CIFAR-10~\cite{krizhevsky2009learning}, CIFAR-100~\cite{krizhevsky2009learning}, Clothing-1M~\cite{Xiao2015Learning} and Webvision-1.0~\cite{Li2017a}. To be fair, we follow most details of the training and evaluation processes from the previous work ``DivideMix''~\cite{li2020dividemix}, such as network architectures, confidence penalty for asymmetric noise, and so on. Our code is publicly available at https://github.com/lijichang/LNL-NCE.

CIFAR-10 and CIFAR-100 are two classic synthetic datasets for the LNL problem. We follow ``DivideMix''~\cite{li2020dividemix} to create the noisy types, \emph{i.e.}, ``Symmetry'' and ``Asymmetry'', and to set noise ratios, namely ``0.20'', ``0.50'', ``0.80'' and ``0.90'' for ``Symmetry'', and ``0.40'' for ``Asymmetry''. Similar to existing works~\cite{liu2020early,li2020dividemix,2021NGC}, we also select PreAct Resnet~\cite{he2016identity} as the model backbone for CIFAR-10/CIFAR-100. Then we train it using a SGD optimizer with a momentum of 0.9 and a weight decay of $5\times10^{-4}$ respectively. To better initialize our model, we set a warm-up step to perform supervised training on the model over all available samples using a standard cross-entropy loss. For effectiveness, this step is assigned a training period $T_{wu}=10$ (or 30) for CIFAR-10 (or CIFAR-100). 
For adapting to diverse scenarios, we empirically set $\tau$ to 0.75 on CIFAR-10 or 0.90 on CIFAR-100, while $\tau^\prime$ are usually set as $2\times10^{-3}$ and $1\times10^{-2}$ on CIFAR-10 and CIFAR-100, respectively.
With respect to other hyper-parameters that are involved in NCE on CIFAR-10/CIFAR-100, we set $K=20, T_{tr}=300,  \gamma=1.0, \eta=0.02$, $B=128$, $B^{\prime}=128$ and $\alpha=4$. 

Clothing-1M and Webvision-1.0 are two large-scale real-world noisy datasets. Clothing-1M contains one million samples grabbed from the online shopping websites and Webvision-1.0 only uses top-50 classes originating from the Google image Subset of Webvision~\cite{Li2017a}. For Webvision-1.0, the results are reported from testing our model on both the WebVision validation set and the ImageNet ILSVRC12 validation set~\cite{szegedy2017inception}.

\begin{table*}[t]
    \caption{Test accuracy (\%) of our method (NCE) and existing state-of-the-art methods on the CIFAR-10 and CIFAR-100 datasets. (Mean accuracy and 95\% confidence interval over 3 trails)}

\scriptsize
  \centering
  \vspace{-0.3cm}
    \begin{tabular}{p{2.3cm}|p{0.8cm}<{\centering}p{0.8cm}<{\centering}p{0.8cm}<{\centering}p{0.8cm}<{\centering}p{1.33cm}<{\centering}|p{0.8cm}<{\centering}p{0.8cm}<{\centering}p{0.8cm}<{\centering}p{0.8cm}<{\centering}}
    \toprule
    \multicolumn{1}{c|}{Dataset } & \multicolumn{5}{c|}{CIFAR-10 } & \multicolumn{4}{c}{CIFAR-100} \\
    \multicolumn{1}{c|}{Noise type } & \multicolumn{4}{c}{Symmetric } & \multicolumn{1}{p{1.33cm}<{\centering}|}{Assymetric } & \multicolumn{4}{c}{Symmetric} \\
    \multicolumn{1}{c|}{Method/Noise ratio } & 0.2   & 0.5   & 0.8   & 0.9   & 0.4   & 0.2   & 0.5   & 0.8   & 0.9 \\
    \midrule
    Cross-Entropy~\cite{li2020dividemix} & 86.8  & 79.4  & 62.9  & 42.7  & 85.0  & 62.0  & 46.7  & 19.9  & 10.1 \\
    F-correction~\cite{patrini2017making} & 86.8  & 79.8  & 63.3  & 42.9  & 87.2  & 61.5  & 46.6  & 19.9  & 10.2 \\
    Co-teaching+~\cite{yu2019coteachingplus}  & 89.5  & 85.7  & 67.4  & 47.9  & -     & 65.6  & 51.8  & 27.9  & 13.7 \\
    PENCIL~\cite{PENCIL_CVPR_2019}  & 92.4  & 89.1  & 77.5  & 58.9  & 88.5  & 69.4  & 57.5  & 31.1  & 15.3 \\
    LossModelling~\cite{arazo2019unsupervised}  & 94.0  & 92.0  & 86.8  & 69.1  & 87.4  & 73.9  & 66.1  & 48.2  & 24.3 \\
    DivideMix~\cite{li2020dividemix}  & {96.1}  & 94.6  & 93.2  & 76.0  & {93.4}  & 77.3  & 74.6  & 60.2  & 31.5 \\
    ELR~\cite{liu2020early}   & 95.8  & {94.8}  & {93.3}  & 78.7  & 93.0  & 77.6  & 73.6  & 60.8  & {33.4} \\
    ProtoMix~\cite{Li2021ICCVLearning}  & 95.8  & 94.3  & 92.4  & 75.0  & 91.9  & 79.1  & 74.8  & 57.7  & 29.3 \\
    NGC~\cite{2021NGC}   & 95.9  & 94.5  & 91.6  & {80.5}  & 90.6  & {79.3} & {75.9} & {62.7}  & 29.8 \\
    \midrule
    \multirow{2}[2]{*}{NCE(best)} & \textbf{96.2} & \textbf{95.3} & \textbf{93.9} & \textbf{88.4} & \textbf{94.5} & \textbf{81.4} & \textbf{76.3} & \textbf{64.7} & \textbf{41.1} \\
          & \tiny{±0.09} & \tiny{±0.12} & \tiny{±0.22} & \tiny{±0.98} & \tiny{±0.70} & \tiny{±0.37} & \tiny{±0.28} & \tiny{±0.56} & \tiny{±0.54} \\
    \midrule
    \multirow{2}[2]{*}{NCE(last)} & 96.0  & 95.2  &  93.6     & 88.0  & 94.2  & 81.0  & 75.3  &  64.5     & 40.7 \\
          & \tiny{±0.22} & \tiny{±0.23} & \tiny{±0.30}     & \tiny{±1.21} & \tiny{±0.96} & \tiny{±0.27} & \tiny{±0.07} & \tiny{±0.86}     & \tiny{±0.42} \\
    \bottomrule
    \end{tabular}%
  \label{tab:cifar}%
\vspace{-0.1cm}
\end{table*}%

\begin{table*}[t]
\scriptsize
    \caption{Test accuracy (\%) of our method (NCE) and existing state-of-the-art methods on the Clothing-1M dataset.}
  \centering
  \vspace{-0.3cm}
  
    \begin{tabular}{cccccc|c}
    \toprule
      Meta-L.~\cite{li2019learning} & DivideMix~\cite{li2020dividemix} & ELR~\cite{liu2020early}   & ELR+~\cite{liu2020early}  & NestedCoT.~\cite{chen2021boosting} & AugDesc~\cite{nishi2021augmentation} &  NCE \\
    \midrule
     73.5  & 74.8  & 72.9  & 74.8  & 74.9  & {75.1}  & \textbf{75.3}\\
    \bottomrule
    \end{tabular}%
    
  \label{tab:Clothing-1M}%
  \vspace{-0.5cm}
\end{table*}%

\begin{table}[t]
\scriptsize
  \caption{Top-1 and top-5 test accuracy (\%) of our method (NCE) and existing state-of-the-art methods on the Webvision and ImageNet ILSVRC12 validation sets. The models are trained on the training set of the Webvision-1.0 dataset}
    \vspace{-0.3cm}
    \centering

    \begin{tabular}{p{2.0cm}|p{1.4cm}<{\centering}p{1.4cm}<{\centering}|p{1.4cm}<{\centering}p{1.4cm}<{\centering}}
    \toprule
    \multicolumn{1}{c|}{\multirow{2}[2]{*}{Method}} & \multicolumn{2}{c}{WebVision} & \multicolumn{2}{|c}{ILSVRC12} \\
            & \multicolumn{1}{p{1.4cm}<{\centering}}{\centering top-1} & \multicolumn{1}{p{1.4cm}<{\centering}|}{\centering top-5} & \multicolumn{1}{p{1.4cm}<{\centering}}{\centering top-1} & \multicolumn{1}{p{1.4cm}<{\centering}}{\centering top-5} \\
    \midrule
    F-correction~\cite{patrini2017making} & 61.1  & 82.7  & 57.4  & 82.4 \\
    
    Decoupling~\cite{malach2017decoupling} & 62.5  & 84.7  & 58.3  & 82.3 \\
    MentorNet~\cite{jiang2018mentornet} & 63.0  & 81.4  & 57.8  & 79.9 \\
    Co-teaching~\cite{han2018coteaching} & 63.6  & 85.2  & 61.5  & 84.7 \\
    DivideMix~\cite{li2020dividemix} & 77.3  & 91.6  & 75.2  & 90.8 \\
    ELR~\cite{liu2020early}   & 76.3  & 91.3  & 68.7  & 87.8 \\
    ELR+~\cite{liu2020early}  & 77.8  & 91.7  & 70.3  & 89.8 \\
    
    NGC~\cite{2021NGC}   & 79.2  & 91.8  & 74.4  & 91.0 \\
    \midrule
     NCE   & \textbf{79.5} & \textbf{93.8} & \textbf{76.3} & \textbf{94.1} \\
    \bottomrule
    \end{tabular}%
\label{tab:webvision}%
\vspace{-0.5cm}
\end{table}

{\bf Baselines.} We compare NCE with the following state-of-the-art algorithms to address the LNL problem on CIFAR-10 and CIFAR-100: ``Cross-Entropy''~\cite{li2020dividemix}, ``F-correction''~\cite{patrini2017making}, ``Co-teaching+''~\cite{yu2019coteachingplus}, ``PENCIL''~\cite{PENCIL_CVPR_2019}, ``LossModelling''~\cite{arazo2019unsupervised}, ``DivideMix''~\cite{li2020dividemix}, ``ELR''~\cite{liu2020early}, ``ProtoMix''~\cite{Li2021ICCVLearning} and ``NGC''~\cite{2021NGC}. Herein, ``Cross-Entropy'' trains the model only with a supervised cross-entropy loss over training samples along with given noisy labels, and its results are copied from ``DivideMix''. Besides methods stated above,  we perform our comparison on Clothing-1M with previous methods, including ``Meta-Learning''~\cite{li2019learning}, ``ELR+''~\cite{liu2020early}, ``NestedCoTeaching''~\cite{chen2021boosting} and ``AugDesc''~\cite{nishi2021augmentation}, where the augmentation strategy of our method on this dataset refers to that of ``AugDesc'' for comparison fairness. Moreover, we evaluate the proposed approach on Webvision-1.0 by newly adding
``Decoupling''~\cite{malach2017decoupling}, ``MentorNet''~\cite{jiang2018mentornet}, and ``Co-teaching''~\cite{han2018coteaching}.


\subsection{Comparisons with the State of the Art}


{\bf Synthetic noisy datasets.} CIFAR-10 and CIFAR-100 are two representative synthetic LNL benchmark datasets and we report results on these datasets in Table~\textcolor{red}{\ref{tab:cifar}}. For fair comparison, we follow all the settings in~\cite{li2020dividemix,2021NGC}. We can see that our NCE outperforms all existing state-of-the-art methods on CIFAR-10 and CIFAR-100 under all settings of symmetric (from 20\% to 90\%) and asymmetric (40\% only) label noise ratio. In particular, on CIFAR-10, our method surpasses the best performing baselines by 7.9\% and 1.1\% at the highest symmetric and asymmetric noise ratios, respectively. In addition, in comparison to the performance of existing algorithms on CIFAR-100, NCE achieves the highest classification accuracy under all four noise ratio settings by exceeding the second best by 2.1\%, 0.4\%, 2.0\% and 7.7\%, respectively.

{\bf Real-world noisy datasets.} To further verify the effectiveness of the proposed NCE method, we also conduct experiments on real-world noisy datasets, namely Clothing-1M and Webvision-1.0. Table~\textcolor{red}{\ref{tab:Clothing-1M}} and Table~\textcolor{red}{\ref{tab:webvision}} show performance comparisons between NCE and existing algorithms when these two are respectively used as the training set. We can observe that NCE achieves the highest accuracy on Clothing-1M and an improvement of 0.2\% over ``AugDesc'', the best performing method among existing ones. Likewise, on the challenging Webvision-1.0, NCE again achieves higher performance than 
most existing methods in terms of top-1 and top-5 accuracy. 
These results further verify that our proposed approach can effectively perform well on the real-world noisy datasets.

\begin{table}[t]
  \centering

\caption{Ablation study of our method (NCE) on the CIFAR-10 and CIFAR-100 datasets under multiple label noise ratios.
``repl.'' is an abbreviation for ``replaced'', and ${\mathcal{L}}^{ce}$ means the model is trained on the clean samples using a cross-entropy loss. (Only one of three trails is selected for comparison in our NCE method)
}
  \vspace{-0.3cm}
    \resizebox{\linewidth}{!}{
    \begin{tabular}{c|l|ccc|cc|c}
    \toprule
    \multirow{3}[2]{*}{M-(\textcolor{red}{\#})} & \multicolumn{1}{c|}{Dataset}   & \multicolumn{3}{c|}{CIFAR-10} & \multicolumn{2}{c|}{CIFAR-100} & \multirow{3}[2]{*}{Mean} \\
          & \multicolumn{1}{c|}{Noise type}  & \multicolumn{2}{c}{Symmetric} & \multicolumn{1}{c|}{Assymetric } & \multicolumn{2}{c|}{Symmetric} \\
          & \multicolumn{1}{c|}{Method/Noise ratio}  & 0.5   & 0.8   & 0.4   & 0.5   & 0.8 \\
    \midrule
    1     & NCE                                                                                                                         & 95.3  & 94.1 & 94.6  & 76.1  & 65.2 & 85.1 \\
    2     & NCE repl. NCNV w/ GMM                                                                                                       & 94.8  & 79.0 & 89.7  & 75.8  & 56.8 & 79.2 \\
    3     & NCE repl. NCLC w/ CT(0.95)                                                                                                  & 94.3  & 86.1 & 90.1  & 76.0  & 58.7 & 81.0 \\
    4     & NCE repl. NCNV w/ GMM \& w/o ${\mathcal{L}}^{lab}$                                        & 91.2  & 78.8 & 87.3  & 71.4  & 49.7 & 75.7 \\
    5     & NCE w/o ${\mathcal{L}}^{lab}$                                                             & 92.5  & 86.7 & 92.6  & 74.4  & 57.9 & 80.8 \\
    6     & NCE repl. ${\mathcal{L}}^{mix}$ w/ ${\mathcal{L}}^{ce}$ & 93.3  & 78.5 & 89.0  & 73.2  & 55.2 & 77.8 \\
    7     & NCE repl. perturbed w/ unperturbed in Eq. (\textcolor{red}{\ref{labloss}})                                                  & 93.6  & 89.4 & 90.5  & 72.5  & 56.1 & 80.4 \\
    
    \bottomrule
    \end{tabular}%
    }
  \label{tab:ablation}%
  \vspace{-0.3cm}
\end{table}%

\subsection{Analysis}

\begin{figure*}[t]
\vspace{-0.3cm}
  \centering
  \resizebox{\linewidth}{!}{
    \begin{tabular}{cccccccccccc}
    \multicolumn{3}{c}{\multirow{6}[0]{*}{\includegraphics[width=4.0cm,height=3.0cm]{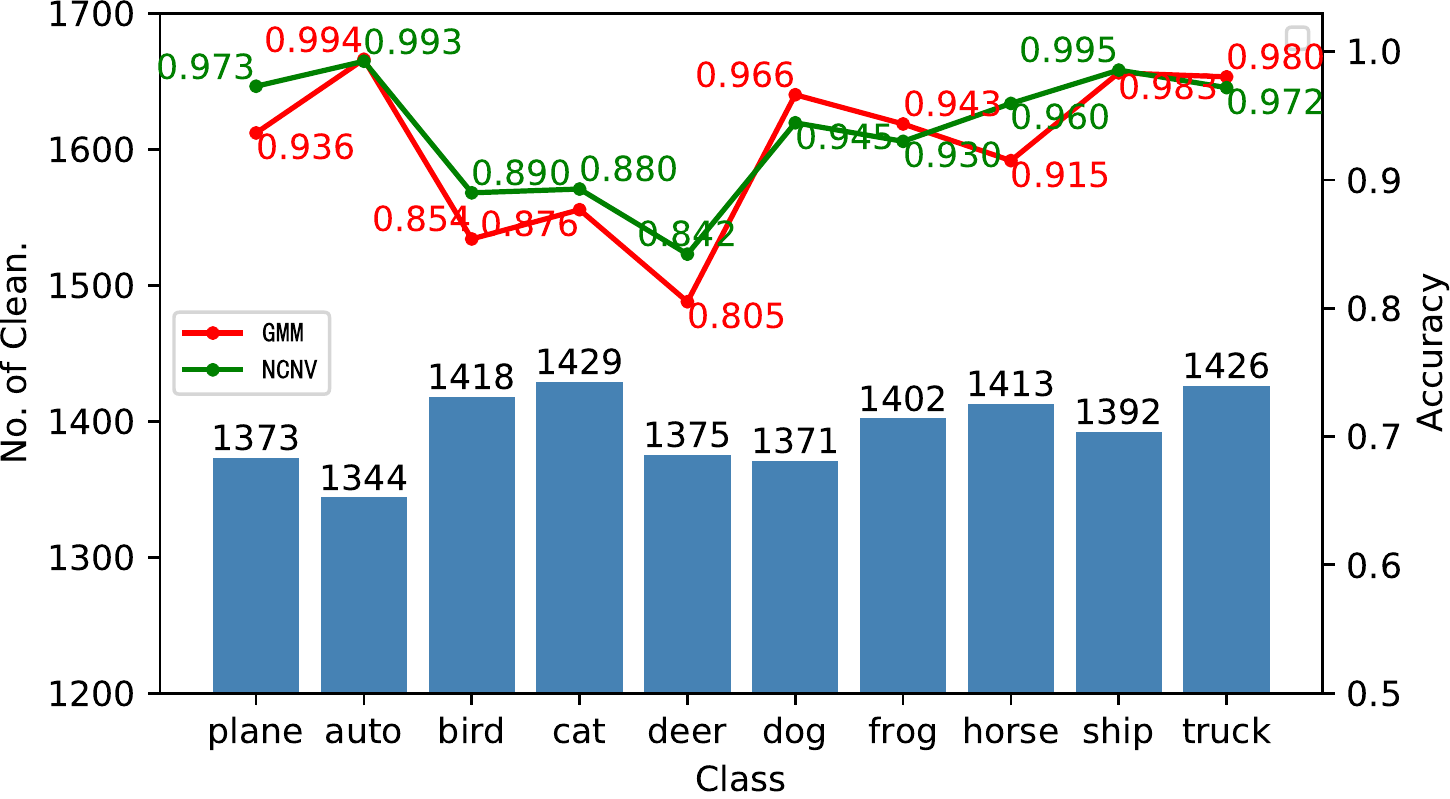}}} & \multicolumn{3}{c}{\multirow{6}[0]{*}{\includegraphics[width=4.0cm,height=3.0cm]{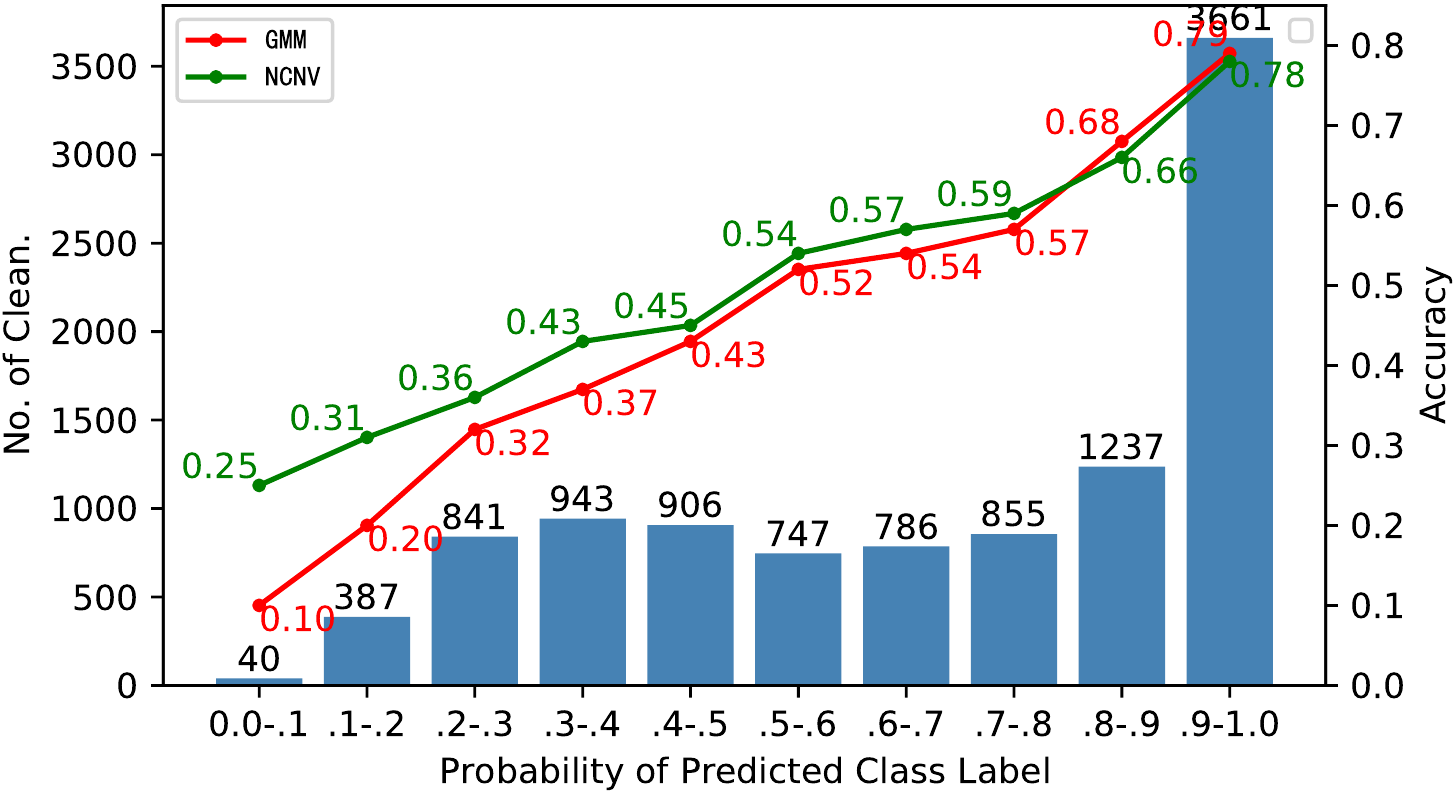}}} &  \multicolumn{3}{c}{\multirow{6}[0]{*}{\includegraphics[width=4.0cm,height=3.0cm]{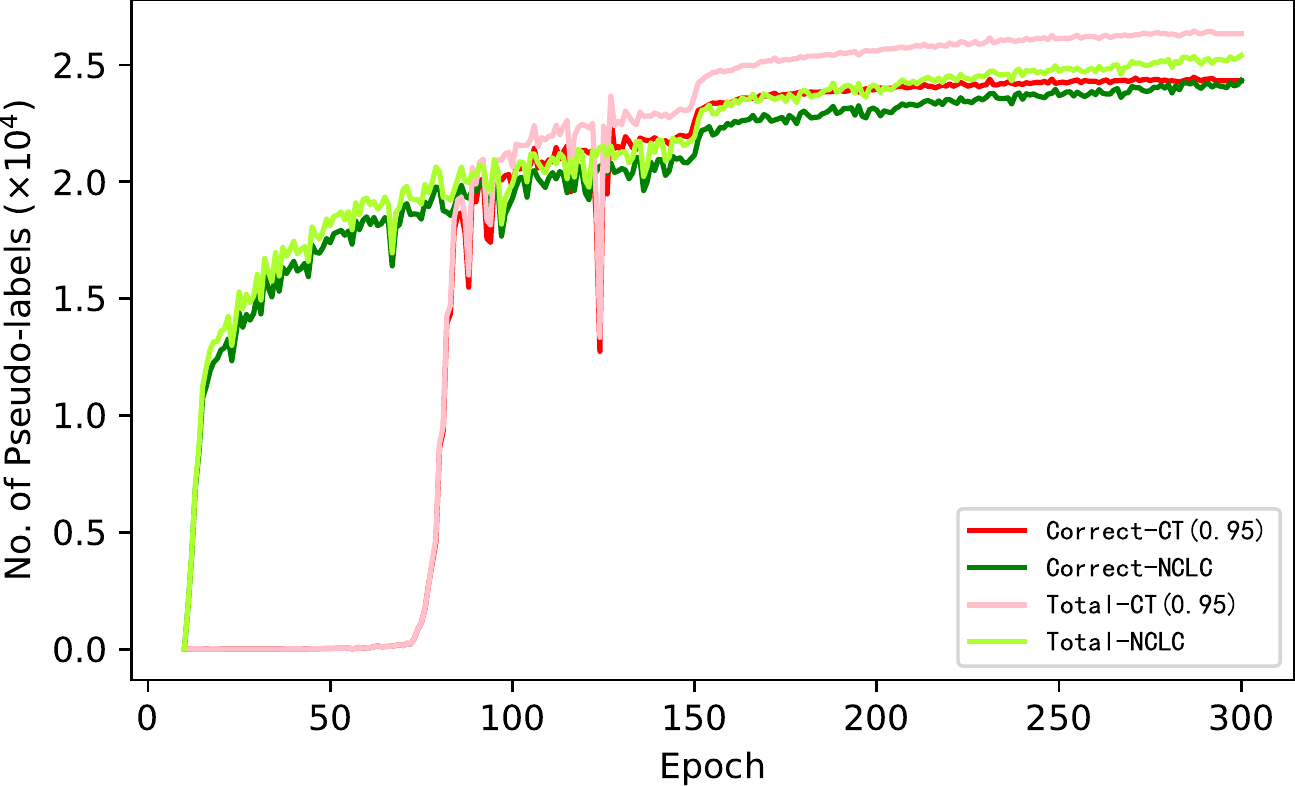}}} & \multicolumn{3}{c}{\multirow{6}[0]{*}{\includegraphics[width=4.0cm,height=3.0cm]{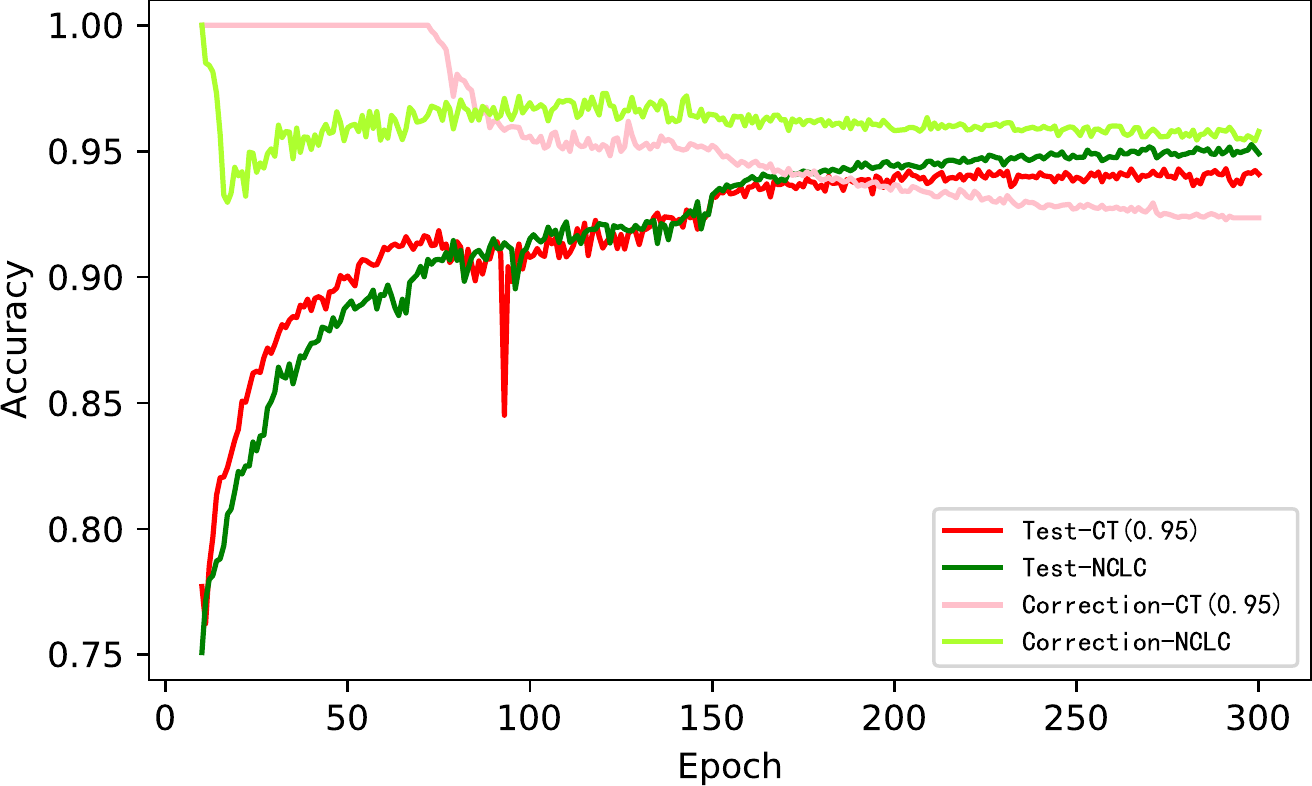}}} \\
    \multicolumn{3}{c}{}  & \multicolumn{3}{c}{}  & \multicolumn{3}{c}{}  & \multicolumn{3}{c}{}  \\
    \multicolumn{3}{c}{}  & \multicolumn{3}{c}{}  & \multicolumn{3}{c}{}  & \multicolumn{3}{c}{}  \\
    \multicolumn{3}{c}{}  & \multicolumn{3}{c}{}  & \multicolumn{3}{c}{}  & \multicolumn{3}{c}{}  \\
    \multicolumn{3}{c}{}  & \multicolumn{3}{c}{}  & \multicolumn{3}{c}{}  & \multicolumn{3}{c}{}  \\
    \multicolumn{3}{c}{}  & \multicolumn{3}{c}{}  & \multicolumn{3}{c}{}  & \multicolumn{3}{c}{}  \\
    \multicolumn{3}{c}{}  & \multicolumn{3}{c}{}  & \multicolumn{3}{c}{}  & \multicolumn{3}{c}{}  \\
    \multicolumn{3}{c}{}  & \multicolumn{3}{c}{}  & \multicolumn{3}{c}{}  & \multicolumn{3}{c}{}  \\
    \multicolumn{3}{c}{{\bf (a)}} & \multicolumn{3}{c}{{\bf (b)}} & \multicolumn{3}{c}{{\bf (c)}} & \multicolumn{3}{c}{{\bf (d)}} \\
    \end{tabular}%
    }
    \vspace{-0.3cm}
    \caption{
    Analysis of ablation study results. {\bf (a)} The accuracy of clean sample identification in various classes.  {\bf (b)} The accuracy of clean sample identification vs. the probability (confidence) of predicted class label.   {\bf (c)} The evolution of the numbers of pseudo-labels and correct pseudo-labels over epochs.  {\bf (d)} The evolution of label correction accuracy and test classification accuracy over epochs. 
     The experiments for (a) and (b) are performed on CIFAR-10 and CIFAR-100 respectively with the same noise profile (Noise ratio: 0.80; Noise type: Symmetric). The blue bars represent the distribution of clean samples. 
    (c) and (d) describe the same experiment, where we analyze the label correction performance of NCLC and Confidence Thresholding (\emph{i.e.}, CT(0.95)) on CIFAR-10 (Noise ratio: 0.50; Noise type: Symmetric)
    }
  \label{fig3}
  
\end{figure*}

\begin{figure}[t]
  \centering
  \resizebox{\linewidth}{!}{
    \begin{tabular}{cccccccccccc}
    \multicolumn{3}{c}{\multirow{6}[0]{*}{\includegraphics[width=3.5cm,height=3.5cm]{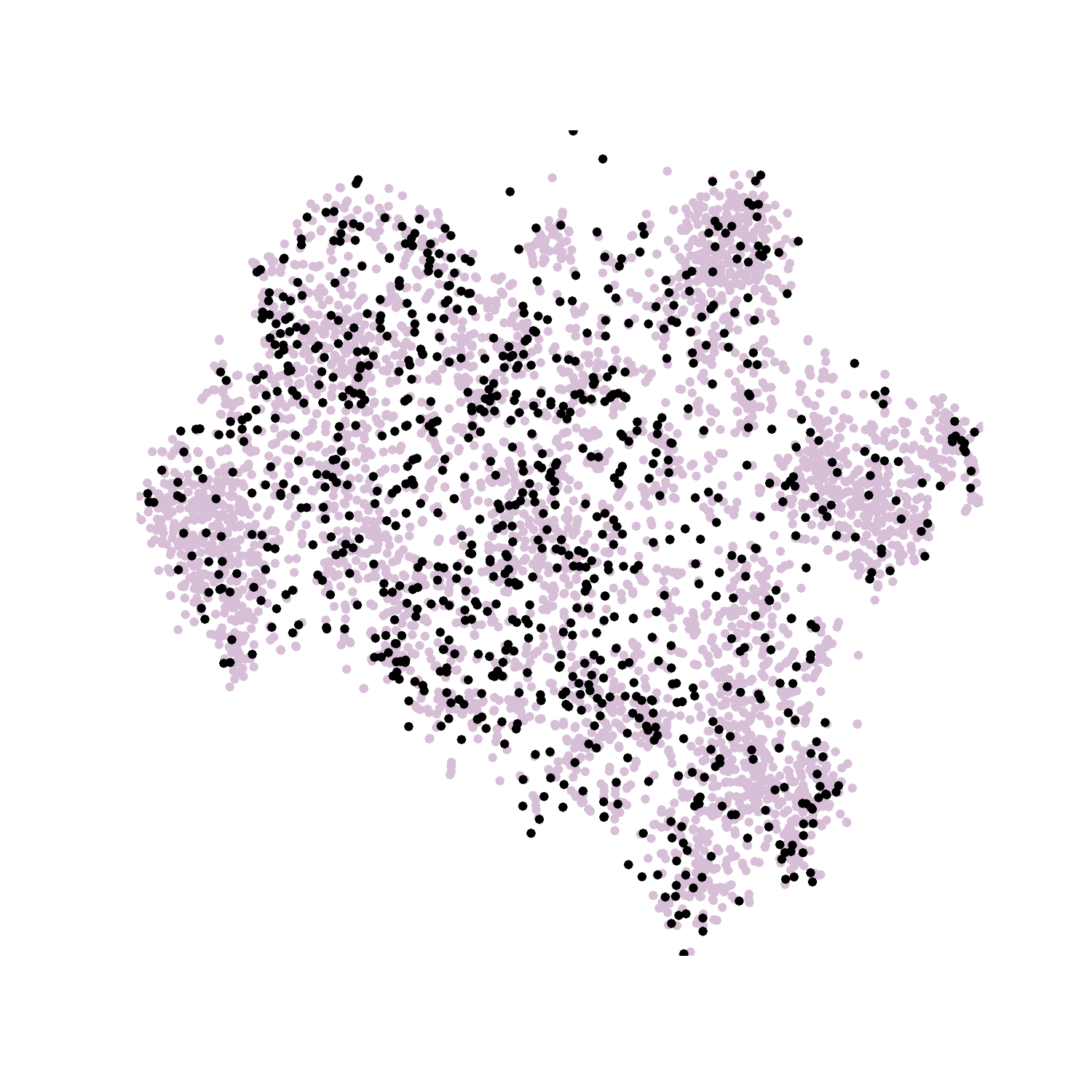}}} & \multicolumn{3}{c}{\multirow{6}[0]{*}{\includegraphics[width=3.5cm,height=3.5cm]{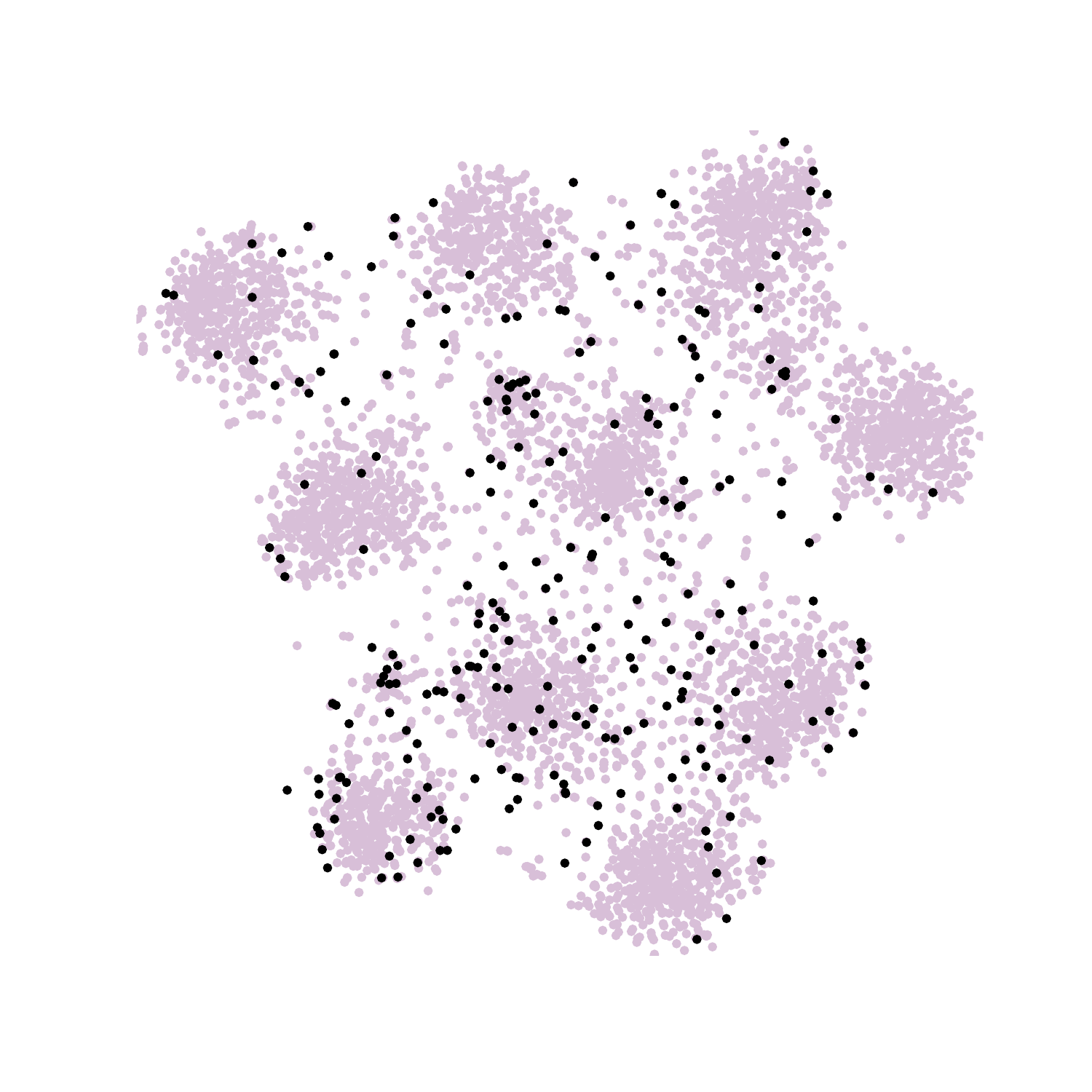}}} & \multicolumn{3}{c}{\multirow{6}[0]{*}{\includegraphics[width=3.5cm,height=3.5cm]{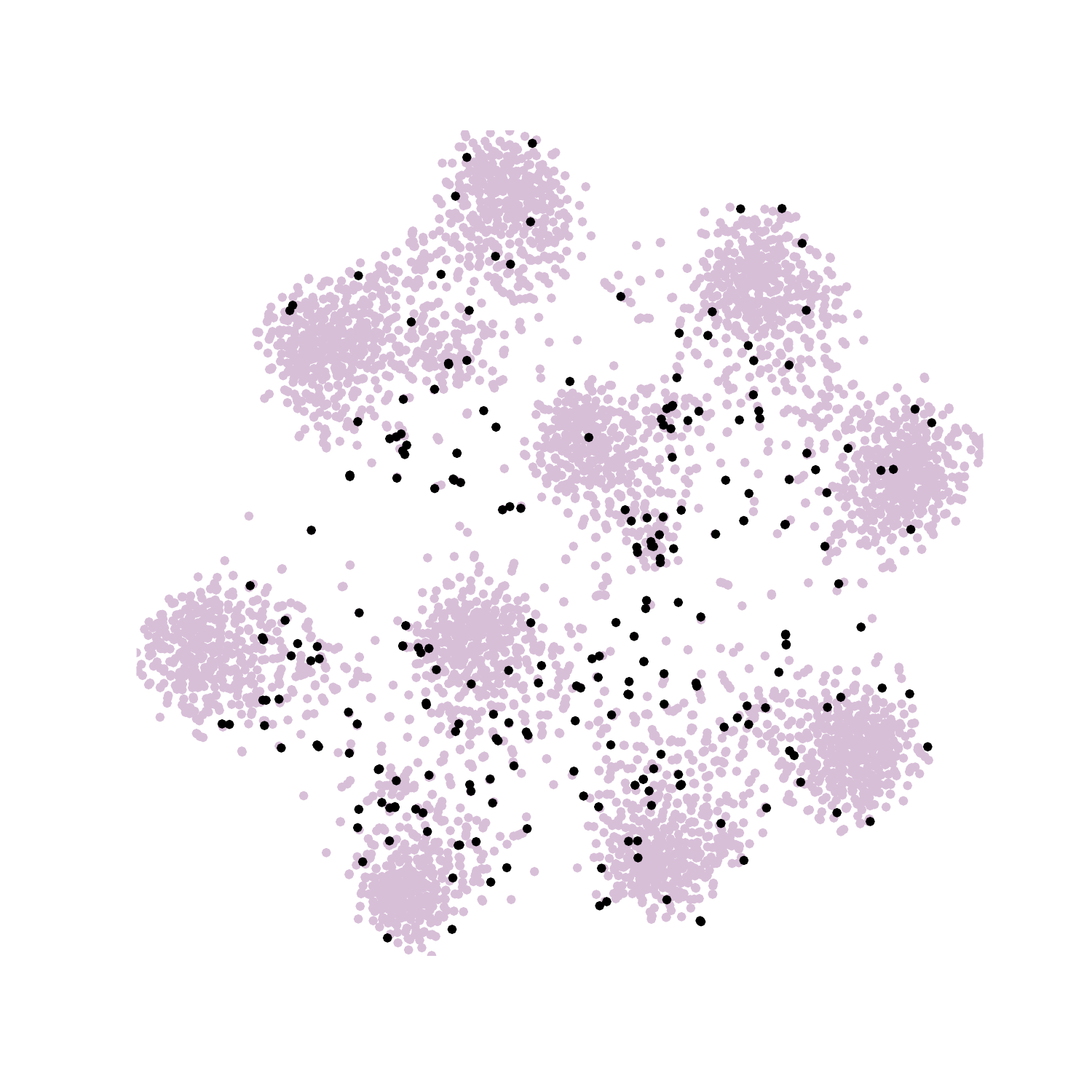}}} & \multicolumn{3}{c}{\multirow{6}[0]{*}{\includegraphics[width=3.5cm,height=3.5cm]{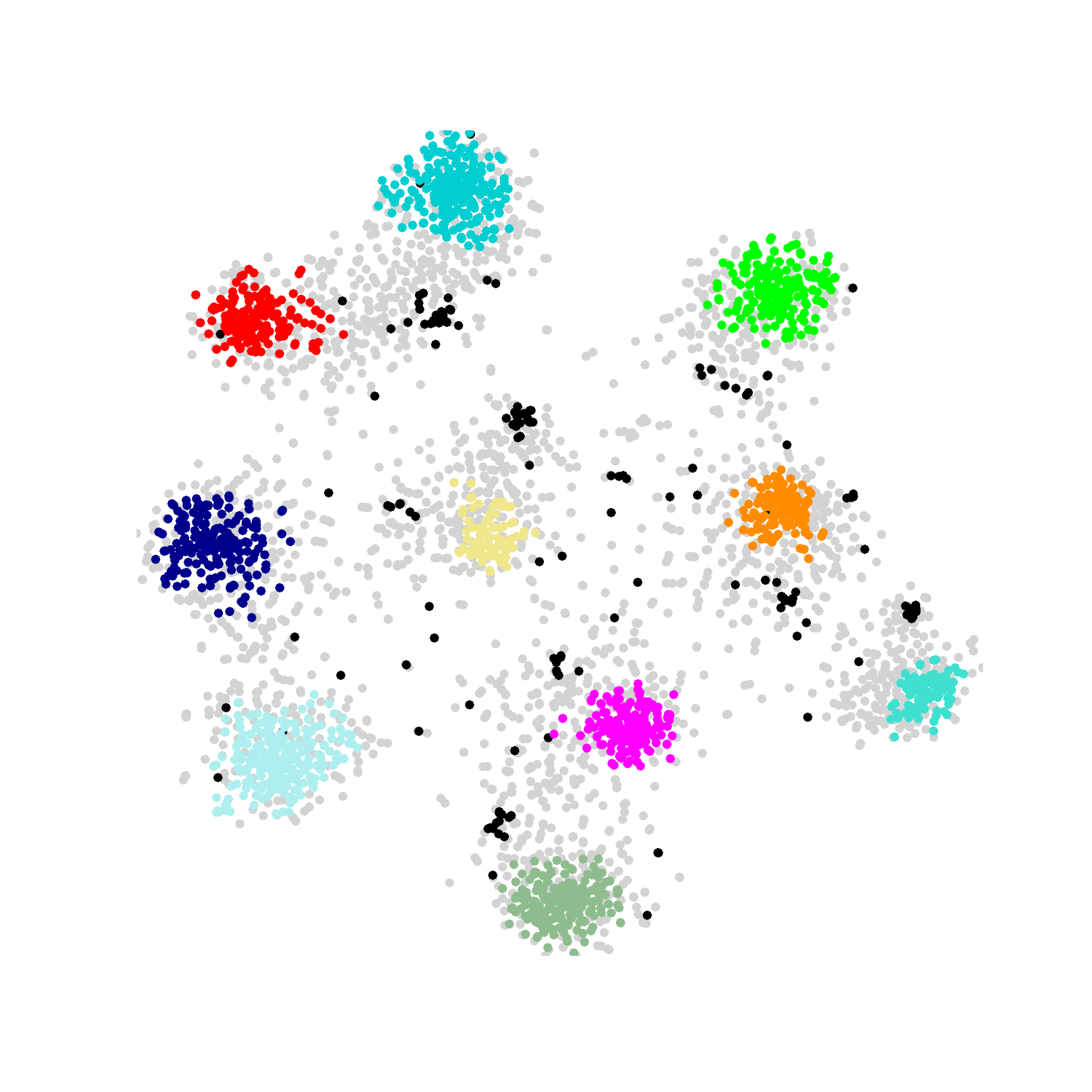}}} \\
    \multicolumn{3}{c}{}  & \multicolumn{3}{c}{}  & \multicolumn{3}{c}{}  & \multicolumn{3}{c}{} \\
    \multicolumn{3}{c}{}  & \multicolumn{3}{c}{}  & \multicolumn{3}{c}{}  & \multicolumn{3}{c}{} \\
    \multicolumn{3}{c}{}  & \multicolumn{3}{c}{}  & \multicolumn{3}{c}{}  & \multicolumn{3}{c}{} \\
    \multicolumn{3}{c}{}  & \multicolumn{3}{c}{}  & \multicolumn{3}{c}{}  & \multicolumn{3}{c}{} \\
    \multicolumn{3}{c}{}  & \multicolumn{3}{c}{}  & \multicolumn{3}{c}{}  & \multicolumn{3}{c}{} \\
    \multicolumn{3}{c}{}  & \multicolumn{3}{c}{}  & \multicolumn{3}{c}{}  & \multicolumn{3}{c}{} \\
    \multicolumn{3}{c}{}  & \multicolumn{3}{c}{}  & \multicolumn{3}{c}{}  & \multicolumn{3}{c}{} \\
    \multicolumn{3}{c}{}  & \multicolumn{3}{c}{}  & \multicolumn{3}{c}{}  & \multicolumn{3}{c}{} \\
    \multicolumn{3}{c}{{\bf (a)} NCNV step, Epoch=30} & \multicolumn{3}{c}{{\bf (b)} NCNV step, Epoch=60} & \multicolumn{3}{c}{{\bf (c)} NCNV step, Epoch=300} & \multicolumn{3}{c}{{\bf (d)} NCLC step, Epoch=300} \\
    \end{tabular}%
    }
    \vspace{-0.3cm}
    \caption{
    Feature visualization using t-SNE. We choose 10 representative classes on CIFAR-100 (Noise ratio: 0.80; Noise type: Symmetric).  (\textbf{a})-(\textbf{c}) show how the distributions of misidentifications in the NCNV step evolve during model training. They are involved in samples from $\mathcal{D}_{\bf train}$ corresponding to each representative class. In these subfigures, points in black are misclassified samples, such as clean (or noisy) samples misclassified as noisy (or clean) ones, in the training data, while samples in purple are correctly identified ones.
    The accuracy of training sample identification in (a)-(c) is 82.2\%, 94.7\% and 95.2\%, respectively.
    (\textbf{d}) shows the feature distributions of unlabeled (noisy) samples in $\mathcal{D}_{{{\bf noisy}}}$ corresponding to 10 classes in the NCLC step, and points in bright colors, black and grey respectively denote correctly relabeled samples, mis-relabeled ones and dropped ones
    }
  \label{fig4}
  \vspace{-0.3cm}
\end{figure}

To provide insights on how effectively each component of our algorithm works, we conduct an ablation study by removing or replacing individual components. Results of this ablation study are summarized in Table~\textcolor{red}{\ref{tab:ablation}} and Fig.~\textcolor{red}{\ref{fig3}}. Also, as displayed in and Fig.~\textcolor{red}{\ref{fig4}}, we perform feature visualization to further analyze the proposed algorithm. All experiments are performed on both CIFAR-10 and CIFAR-100 datasets.

{\bf Effectiveness of NCNV step.} To examine the effectiveness of the NCNV step in identifying clean/noisy labels, we replace NCNV with a well-known GMM-based strategy proposed in ``DivideMix''~\cite{li2020dividemix}. In Table~\textcolor{red}{\ref{tab:ablation}}, a comparison between row M-(\textcolor{red}{1}) and row M-(\textcolor{red}{2}) reveals that our NCNV step significantly outperforms the GMM-based strategy because the former is capable of identifying clean labels of harder samples. Specifically, Fig.~\textcolor{red}{\ref{fig3}}(a) and (b) show the power of our NCNV step in handling ``hard'' classes and ``hard'' samples in the clean subset. A class is considered ``hard'' when multiple methods have an overall low clean sample identification accuracy in the class, while a ``hard'' sample has a low probability (confidence) associated with its predicted class label. As Fig.~\textcolor{red}{\ref{fig3}}(a) shows, our method achieves higher sensitivity on ``hard" classes, \emph{i.e.} ``cat'', ``bird'' and ``deer'', where both methods have the lowest identification accuracy.
In addition, Fig.~\textcolor{red}{\ref{fig3}}(b) also shows that our NCNV step works significantly better on ``hard" samples, whose predicted class labels are associated with a low probability (confidence).

{\bf Effectiveness of NCLC step.} To better understand the performance of the NCLC step in label correction, we replace NCLC with an existing label correction scheme, called Confidence Thresholding (CT)~\cite{sohn2020fixmatch}, which relabels such samples whose pseudo-labels have a confidence value exceeding a predefined threshold, \emph{e.g.}, 0.95. According to row M-(\textcolor{red}{3}) of Table~\textcolor{red}{\ref{tab:ablation}}, NCLC clearly outperforms CT under all noise ratio settings. In detail, Fig.~\textcolor{red}{\ref{fig3}}(c) and (d) reveal that CT works with few pseudo-labels in the early epochs. This is because, at that moment, the model cannot fit the training samples well and thus unlabeled samples with low-confidence predictions ($<0.95$) would not be assigned pseudo-labels. Afterwards, although plenty of unlabeled samples are given pseudo-labels as model training goes on, the label correction accuracy drops at the same time. Ultimately, it leads to lower performance than NCLC, which, on the other hand, obtains more reliable pseudo-labels for unlabeled (noisy) points.

{\bf Necessity of mixup regularization.} To verify the importance of the mixup regularization, we remove it from our algorithm and then perform standard supervision over clean samples. As shown in row M-(\textcolor{red}{6}) of Table~\textcolor{red}{\ref{tab:ablation}}, this change causes very serious performance degradation, indicating that the mixup regularization is able to effectively attenuate noise memorization.

{\bf Necessity of consistency regularization.}
To investigate the effectiveness of consistency regularization over unlabeled (noisy) samples, we conduct two experiments. First, we disable ${\mathcal{L}}^{lab}$, meaning that the model is only trained over all clean samples. By comparing row M-(\textcolor{red}{5}) with row M-(\textcolor{red}{1}) in Table~\textcolor{red}{\ref{tab:ablation}}, we observe that the performance under all noise ratios drops by $1.7\%$ to $7.6\%$, suggesting that this consistency loss is important for the performance of the model, especially when the noise ratio is high. In the second experiment, we replace the perturbed samples used in Eq. (\textcolor{red}{\ref{labloss}}) with unperturbed ones to examine the need of sample perturbations. As shown in row M-(\textcolor{red}{7}) of Table~\textcolor{red}{\ref{tab:ablation}}, the average accuracy drops considerably by $4.7\%$. 
This demonstrates that sample perturbations in Eq. (\textcolor{red}{\ref{labloss}}) play a significant role in realizing the full potential of consistency regularization.
 
{\bf Feature visualization.} We use t-SNE~\cite{van2008visualizing} to visualize the feature distributions in both NCNV and NCLC steps. In Fig.~\textcolor{red}{\ref{fig4}}(a)-(c), we show how the distributions of misidentified samples across diverse classes evolve in the model training process. It can be observed that as model training proceeds, the number of misclassifications in the training data decreases gradually. The misclassifications are distributed near the boundaries of the clusters corresponding to the classes, showing a good noise verification effect. Furthermore, in the NVLC step, as illustrated in Fig.~\textcolor{red}{\ref{fig4}}(d), most well-relabeled samples are located in the core regions of the clusters, while the mis-relabeled points and dropped ones are closer to the boundaries of the clusters or peripheral areas between different clusters. This meets our assumption stated in Section (\textcolor{red}{3.2}) that a candidate sample in the NVLC step that satisfies Eq. (\textcolor{red}{9}) is more likely to be farther away from the decision boundary of the model and could derive a more reliable pseudo-label.
\vspace{-0.2cm}
\section{Conclusions}
\vspace{-0.1cm}
In this paper, we have introduced a novel method called Neighborhood Collective Estimation (NCE) to tackle the problem of learning with noisy labels. In this method, we re-estimate the predictive reliability of a candidate sample by contrasting it against its feature-space nearest neighbors. This can enrich and diversify predictive information associated with the candidate and also makes such information relatively unbiased. The accuracy of noisy label identification and correction can thus be improved, facilitating subsequent model training. In detail, NCE consists of two steps, 1) Neighborhood Collective Noise Verification (NCNV) for separating all training data into clean samples and noisy ones, and 2) Neighborhood Collective Label Correction (NCLC) for relabeling noisy samples. Extensive experiments and a thorough ablation study have confirmed the superiority of our proposed method.

\vspace{-0.2cm}
\section*{Acknowledgements}
\vspace{-0.15cm}
This work was supported in part by the Guangdong Basic and Applied Basic Research Foundation (No.2020B1515020048), in part by the National Natural Science Foundation of China (No.61976250, No.U1811463), in part by the Guangzhou Science and technology project (No.202102020633), and in part by Hong Kong Research Grants Council through Research Impact Fund (Grant R-5001-18).

\clearpage
%
%
\bibliographystyle{splncs04}
\bibliography{manuscript}
\end{document}